\documentclass[cameraready]{Interspeech}
\usepackage{graphicx}
\usepackage{multirow}
\usepackage{multicol}
\usepackage{subcaption}
\usepackage{amsmath}
\usepackage{amssymb}
\title{Localizing and Editing Knowledge in Large Audio-Language Models}

\author[affiliation={1}]{Sung Kyun}{Chung}
\author[affiliation={1}, equalcontribution]{Jiaheng}{Dong}
\author[affiliation={1}, equalcontribution]{Qiuchi}{Hu}
\author[affiliation={3}]{Gongping}{Huang}
\author[affiliation={2}]{Hong}{Jia}
\author[affiliation={1}]{Ting}{Dang}

\address{
     $^1$ The University of Melbourne, Australia \\
    $^2$ University of Auckland, New Zealand \\
    $^3$ Wuhan University, China
}

\email{sungkyun.chung1@student.unimelb.edu.au, jiahengd@student.unimelb.edu.au, qiuchi.hu@student.unimelb.edu.au}


\usepackage{comment}
\usepackage{soul, xcolor}

\begin{document}

\maketitle

\begin{abstract}
Large Audio–Language Models (LALMs) have shown strong performance in speech understanding, making speech a natural interface for accessing factual information. Yet they are trained on static corpora and may encode incorrect facts. Existing model editing methods localize and update facts in text-only LLMs, but do not account for continuous speech representations, or where knowledge is stored across audio encoder or LLM backbone in LALMs. 
We construct the first audio benchmark for knowledge localization and editing in LALMs and propose a speech-driven locate–then–edit framework. First, we use speech-aware causal tracing to localize layers 
that support factual retrieval and then apply editing at identified sites. Experiments show that factual knowledge is jointly encoded in layers from audio encoder and language backbone, 
and that audio editing yield more effective updates than text editing or fine-tuning, enabling fine-grained knowledge control in speech AI systems.

\end{abstract}

\vspace{-5pt}
\section{Introduction} 
Large Audio-Language Models (LALMs) have recently demonstrated strong capabilities in speech understanding, instruction following, audio question answering, and spoken interaction~\cite{chu2024qwen2audiotechnicalreport,Das2024SpeechVerseAL,tang2024salmonngenerichearingabilities, goel2025audio, ghosh-etal-2024-gama}. By jointly modeling acoustic and linguistic signals, these systems can reason over spoken inputs and generate contextually appropriate responses.
Despite their impressive performance, LALMs are trained on large but static corpora. They may encode outdated, incomplete, or incorrect factual associations, leading to potential hallucinations and unreliable outputs~\cite{leng2024cursemultimodalitiesevaluatinghallucinations,kuan2024understandingsoundsmissingquestions, kuan2024largeaudiolanguagemodelstruly}. However, refreshing model knowledge through supervised fine-tuning or continual learning is computationally expensive and often infeasible at scale. 

Unlike text-only LLMs, which operate purely on discrete tokens, LALMs incorporate dedicated 
audio encoders and 
further post-train the LLM backbone to jointly process multimodal representations from both audio and text
~\cite{chu2024qwen2audiotechnicalreport, Das2024SpeechVerseAL, tang2024salmonngenerichearingabilities}. Prior work on text LLMs shows that factual associations can often be localized to specific 
transformer layers or to specific modules within transformer layers, such as attention or MLP components, 
and updated via targeted parameter modifications~\cite{rome, memit, alphaedit}. However, these findings assume a language-centric storage pattern and do not account for modality-specific encoders or cross-modal interactions. In LALMs, factual knowledge may instead be distributed across 
layers or modules in audio encoder
, language backbone, 
and their interfaces, raising new questions about where facts are stored and how edits propagate through the multimodal computation.

Existing approaches to improving factual reliability largely fall into two categories: retrieval-based methods (e.g., RAG)~\cite{DBLP:journals/corr/abs-2005-11401, DBLP:journals/corr/abs-1911-00172} that inject external knowledge, and model editing methods that directly modify internal parameters\cite{rome,memit,alphaedit}. Retrieval methods can flexibly incorporate up-to-date information but incur additional storage and runtime costs and require careful retrieval and integration. Model editing is more lightweight, as it updates only a small subset of weights so that the desired fact is reflected in the model’s predictions without full retraining. Nevertheless, all widely used editing methods (e.g., ROME~\cite{rome}, MEMIT~\cite{memit}, AlphaEdit~\cite{alphaedit}) have been developed for text-only LLMs, leaving it unclear whether they can reliably locate and update factual knowledge in LALMs without disrupting audio–text alignment or multimodal reasoning.

Compared to text-based LLMs, standard model editing methods, i.e., locate–then–edit methods, typically rely on controlled corruption of discrete input tokens (e.g., perturbing subject tokens) to trace causal pathways and localize factual knowledge before editing~\cite{rome,memit,alphaedit}. Extending this strategy to LALMs is non-trivial. Text consists of tokens with clear semantic boundaries, whereas speech is a continuous waveform with temporally distributed representations~\cite{yue2023token2vec}. While text corruption can use well-defined token substitutions, speech corruption must perturb continuous acoustic signals over time, making it difficult to define token-like boundaries in the acoustic space. Moreover, attribution in LALMs must go beyond text-level semantics: effective localization and editing require understanding not only semantic subspaces but also how content is distributed across temporally structured and multimodal representations.

To address these gaps, we (i) curate, to our knowledge, the first audio benchmark for knowledge localization and editing in LALMs; (ii) propose a speech-driven locate–then–edit framework that combines causal tracing with multimodal model editing; and (iii) conduct a systematic study across two datasets. Our framework first uses speech-aware causal tracing to identify layers and modules that contribute to factual retrieval from spoken prompts, and then applies single-layer and multi-layer editing at identified multi-modal locations. Experiments show that factual knowledge is jointly encoded in transformer layers from both audio encoder and language backbone
, with layers from audio encoder 
playing a more prominent role. Coordinated cross-modal edits yield more effective updates than unimodal editing or standard fine-tuning. These findings provide initial evidence for principled, fine-grained control of factual knowledge in speech-based AI systems.

\begin{figure*}[t]
    \centering
    \includegraphics[width=0.9\textwidth]{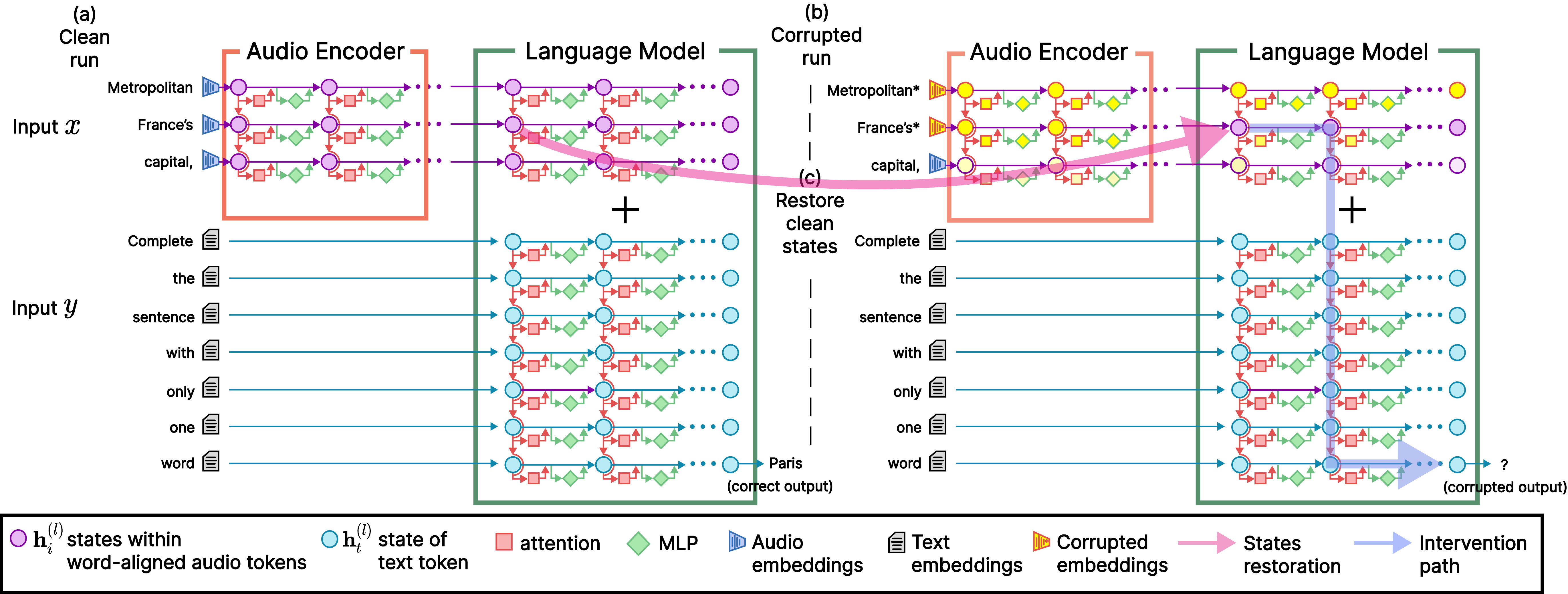}
    \caption{Causal tracing consists of three runs: i) a clean run for the correct object; ii) a corrupted run, where the subject span are corrupted to obtain a low-probability object prediction; and iii) a corrupted-with-restoration run, which restores the corrupted hidden representations at selected layers to their clean values, evaluating the contribution of those layers via changes in object probability.}
    \vspace{-15pt}
    \label{fig:overall}
\end{figure*}

\vspace{-5pt}
\section{Related Work}
\noindent\textbf{Large Audio–Language Models.}
Current LALMs extend text-based LLMs with audio encoders 
that map raw audio into token-compatible representations, while the LLM backbone is post-trained 
for joint audio–text reasoning 
(e.g., BLSP~\cite{blsp}, SALMONN~\cite{tang2024salmonngenerichearingabilities}, Qwen2-Audio~\cite{chu2024qwen2audiotechnicalreport}). These architectures support diverse speech and audio tasks, but modality-specific encoders and cross-modal interactions may substantially change how factual knowledge is represented compared to text-only LLMs.

\noindent\textbf{Knowledge Localization via Causal Tracing.}
Causal tracing identifies where information is stored in LLM transformer layers 
by measuring how interventions on internal representations affect predictions~\cite{rome}. It typically corrupts hidden states and selectively restores them during inference to reveal components responsible for correct outputs. Early work showed that feed-forward (FFN) layers behave as key–value memories for factual associations~\cite{mlpkeyvalue}, and later studies found that localization patterns depend on architecture, modality, and training regime~\cite{rome, visionLLM}. However, existing analyses focus on text-only LLMs; analogous investigations for LALMs remain largely unexplored.

\noindent\textbf{Model Editing.}
Model editing methods either preserve parameters or directly modify them. Parameter-preserving approaches such as MELO~\cite{yu2023meloenhancingmodelediting} and in-context knowledge editing (IKE)~\cite{zheng-etal-2023-edit} externalize new knowledge via auxiliary mechanisms, avoiding weight changes but adding storage or context overhead and keeping edits implicit. Parameter-modifying methods update a subset of weights so that a target factual association is reflected in the model output. Locate–then–edit methods like ROME~\cite{rome}, MEMIT~\cite{memit}, and AlphaEdit~\cite{alphaedit} apply targeted updates at single or multiple transformer 
layers to inject or revise facts. However, all of these techniques are developed for text-only LLMs. In LALMs, where knowledge may be jointly represented across audio encoder and LLM backbone, 
it is unclear whether such unimodal strategies can reliably locate and edit distributed factual knowledge without degrading audio–text alignment or multimodal reasoning.

\vspace{-0.5em}
\section{Method}
We study how factual knowledge is stored and updated in LALMs with spoken input. We represent each fact as a triple $(s, r, o^{c})$ in dataset $D$, where \(s\) denotes the subject, \(r\) the relation, and \(o^c\) the object currently stored in the model. Specifically, to probe a fact in the spoken modality, we construct a speech prompt \(x = T(s,r)\) (e.g., $x = $"Metropolitan France's capital,", $s = $ “Metropolitan France's”, $r = $ “capital,”) in audio and provide a fixed textual instruction $y$ that asks the model to complete the prompt. The predicted probability of the model for the object token $o^{c}$ (e.g., $o^{c} = $“Paris”) then reflects how strongly that factual association is stored in the LALM.

Our method follows a two-stage \emph{locate–then–edit} framework. In the \emph{locate} stage, we apply causal tracing to quantify the contribution of each transformer layers and attention/MLP modules within these layers 
in the LALM. 
In the \emph{edit} stage, we investigate multiple editing methods to the layers or modules identified in the locate stage, modifying their parameters so that the model’s prediction shifts from the original object (e.g., “Paris”) to a new object (e.g., “Lyon”), while aiming to preserve unrelated knowledge.

\vspace{-5pt}
\subsection{Locate via Causal Tracing}

As shown in Figure~\ref{fig:overall}, we use causal tracing to identify which internal representations of the LALM store factual knowledge. Each procedure has three runs: a clean run, a corrupted run, and a corrupted-with-restoration run. 

\vspace{2pt}
\noindent \textbf{Clean Run. }
Given an input pair $(x,y)$ with speech $x$ and prompt $y$, we feed it to the LALM and record each transformer layer output representations as 
hidden states at every frame $i$ and layer $\ell$ as ${h_i^{(\ell)} \mid i \in [1,T], \ell \in \mathcal{L}}$ (Figure~\ref{fig:overall}(a)). The probability of the correct object token $P(o^c \mid x,y)$ serves as the \emph{high-score} baseline.


\vspace{2pt}
\noindent \textbf{Corrupted Run. }
To disrupt the factual association, we corrupt the speech representations aligned with the subject span. A key challenge in speech corruption lies in accurately identifying subject phrase boundaries within continuous audio. We obtain word-level timestamps using WhisperX forced alignment~\cite{whisperx}, map them to acoustic frames, and corrupt only those frames (e.g., “Metropolitan France’s”). The corruption is applied to the input representations to the first transformer layers of the audio encoder
, where low-level acoustic features remain stable but still encode subject identity. We inject Gaussian noise into the subject-span:
\begin{equation}
h_i^{*,(0)} \leftarrow h_i^{(0)} + \epsilon,
\quad \epsilon \sim \mathcal{N}(0, 3\sigma).
\end{equation}
where $\sigma$ is the standard deviation of the clean subject-span representations following~\cite{rome}.  
This yields corrupted states $h_{i}^{*,(\ell)}$ and a low object probability $P^{*}(o^c \mid x,y)$, used as the \emph{low-score} baseline.



\vspace{2pt}
\noindent \textbf{Corrupted-with-restoration Run.}
To measure the contribution of a specific layer, we restore clean activations during the corrupted run. Specifically, we start from the corrupted input and run forward to frame $\hat{i}$ and layer $\hat{\ell}$, then restore a small layer window around $\hat{\ell}$ to their clean values:
\begin{equation}
    h_{\hat{i}}^{(\ell)} \leftarrow  h_{\hat{i}}^{*,(\ell)},
\quad \ell \in [\hat{\ell}-4,\,\hat{\ell}+4],
\end{equation}
We adopt $\pm 4$ layer window to allow the restored signal to propagate reliably, while preserving layer-level attribution as~\cite{rome}.

The model continues the forward pass to produce $P^{\mathrm{restore}}(o^c \mid x,y)$. The \emph{Indirect Effect} (IE) of layer $\hat{\ell}$ is:
\begin{equation}
\mathrm{IE}^{(\hat{\ell})} = P^{*}_{\mathrm{restore}}(o^c \mid x,y) - P^{*}(o^c \mid x,y).
\end{equation}
and the Average IE (AIE) is obtained by averaging over all speech sentences. Transformer 
layers with high AIE $\mathrm{AIE}^{(\hat{\ell})}$ are treated as primary storage sites for the association between $s$ and $o^c$ and are used as targets for subsequent model editing. We further decompose causal tracing by separately intervening on the MLP and attention modules 
within each layer, across both the audio encoder and LLM backbone, for multimodal analysis. 

\vspace{-5pt}
\subsection{Model Editing}
Given an existing fact $(s, r, o^c)$, model editing aims to update the LALM so that it instead reflects a target fact $(s, r, o_{\text{target}})$ while preserving unrelated behaviour. For brevity, we omit layer and frame indices. Following prior work on text LLMs~\cite{rome}, we assume a linear transformation:
\begin{equation}
    h = Wk, 
\end{equation}
where $k = \phi(s,r)$ is the hidden representation (key) induced by the subject–relation input and $W$ is a learned weight matrix. The current output $Wk = h^c$ corresponds to the representation of $o^c$. Editing amounts to finding a perturbation $\Delta$ such that:
\begin{equation}
   (W+ \Delta)k = v^{\text{target}}, 
\end{equation}
We study four editing strategies with increasing flexibility.

\vspace{2pt}
\noindent\textbf{Single-Layer Editing.}
Single-layer editing applies a constrained rank-one update. Let the residual
\begin{equation} \label{eq:R_formula}
R = v^{\text{target}} - Wk,
\end{equation}
be the shift required to align the current output with the target. Single-layer editing computes
\begin{equation}\label{eq:rome}
\Delta = R k^{T} (k k^{T} + \lambda I)^{-1},
\end{equation}
which minimises the update magnitude while enforcing $(W+\Delta)k \approx v^{\text{target}}$. Unlike text-only LLMs, audio words span multiple frames, so we obtain word-level representations via mean pooling over aligned audio frames and use the first or last subject word as $k$. We apply this procedure to (i) a single audio layer and (ii) a single text layer respectively. 


\vspace{2pt}
\noindent\textbf{Sequential Cross-Modal Single-Layer Editing.}\label{sec:seq}
We extend single-layer editing to a sequential audio–text setting. We first edit a selected audio encoder layer, and the resulting representation from the audio encoder performs a subsequent update in LLM backbone layers through the forward pass. 
In this way, language-path edits are explicitly conditioned on the audio edit, ensuring that semantic changes remain grounded in the updated acoustic representation. 
%



\vspace{2pt}
\noindent\textbf{Multi-Layer Editing.}
Rank-one updates at a single layer can unintentionally affect other facts with overlapping keys. To better control interference, we perform multi-layer editing with explicit preservation constraints. Let $K_0$ be keys of protected facts, $K_1$ keys to be edited, and $K_p$ keys from previously edited facts. We require
\vspace{-5pt}
\begin{equation}
(W + \Delta)K_0 = WK_0,
\end{equation}
which is enforced by projecting updates into the null space of $K_0$. Let $P$ be a projection matrix with $PK_0=0$. We solve
\begin{align}
\Delta = \arg\min_{\widetilde{\Delta}} \Big( & \lVert (W + \widetilde{\Delta}P)K_1 - V_1 \rVert^2 \\ \nonumber
& + \lVert (W + \widetilde{\Delta}P)K_p - V_p \rVert^2 \nonumber
& + \lVert \widetilde{\Delta}P \rVert^2 \Big).
\end{align}
and apply the projected update $\Delta_{\text{multi}} = \Delta P$ to selected layers. Following~\cite{alphaedit}, we distribute a single edit direction across multiple layers using predetermined layer-wise weights. We apply the multi-layer editing to i) audio or ii) text layers respectively.

\noindent\textbf{Sequential Cross-Modal Multi-Layer Editing.} 
Finally, we perform sequential multi-layer editing across both audio encoder and LLM backbone. 
We first apply $\Delta_{\text{multi}}$ to target layers in $E_{\text{audio}}$, then edit layers in $F_{\text{LLM}}$ conditioned on the modified audio encoder. This cross-modal variant offers greater flexibility for updating distributed factual representations while maintaining alignment between audio and text pathways.

\vspace{-5pt}
\section{Experimental Setup}

\textbf{Dataset. }
We use CounterFact \cite{rome} and Known-1000~\cite{rome} in our experiments.
CounterFact is a text-based benchmark of factual triples $(s, r, o^c)$. 
Known-1000 consists of factual prompts correctly completed by GPT-2 XL. 
We first generate a spoken realization of this dataset using the Gemini 2.5 Flash TTS model~\cite{gemini25_tts},
obtaining full audio sentences consisting of the subject, relation and object. 


For localization, we retain only instances for which the pre-edit LALM predicts the correct object $o^c$, resulting in 250 Known-1000 and 100 CounterFact examples. For editing, we construct a 500-instance CounterFact subset comprising these 200 filtered items plus 300 additional randomly sampled triples without any pre-edit constraint.

\vspace{2pt}
\noindent\textbf{Implementation Details.}
We use Qwen2-Audio-7B-Instruct as the backbone LALM for all causal tracing and editing experiments. For single-layer and sequential cross-modal editing, we optimize for 10 gradient steps with learning rate $0.01$. Audio-layer edits use a weight decay of $0.5$, while text-layer edits use no weight decay. For the multi-layer setting, we compute protected keys $K_0$ using 5000 samples drawn from LibriSpeech train-clean-100 \cite{panayotov2015librispeech}. 
We optimize for 10 gradient steps with learning rate $0.5$ and weight decay of $0.5$ for audio- and text-layer edits. 
As an additional baseline, we perform lightweight fine-tuning of selected layers using Adam with learning rate $5\times 10^{-4}$. All experiments run on a single NVIDIA A100 GPU.

\vspace{2pt}
\noindent\textbf{Evaluation. }
We adopt the standard evaluation protocol~\cite{rome} and report three metrics. The \emph{Efficacy Score} (ES) measures whether the edited model prefers the target object $o_{\text{target}}$ over the original object $o^c$ after editing. The \emph{Paraphrase Score} (PS) evaluates generalization to paraphrased prompts of the same $(s,r)$ pair. The \emph{Neighbor Score} (NS) measures specificity by testing whether related triples sharing relation $r$ still preserve the original object $o^c$ after editing. 
High ES and PS indicate effective and robust edits, while high NS reflects minimal unintended changes. We additionally report an overall score $S$, computed as the mean of ES, PS, and NS. Details can be referred to~\cite{rome}.

\begin{figure}[t]
    \centering
    
    \includegraphics[width=\columnwidth]
    {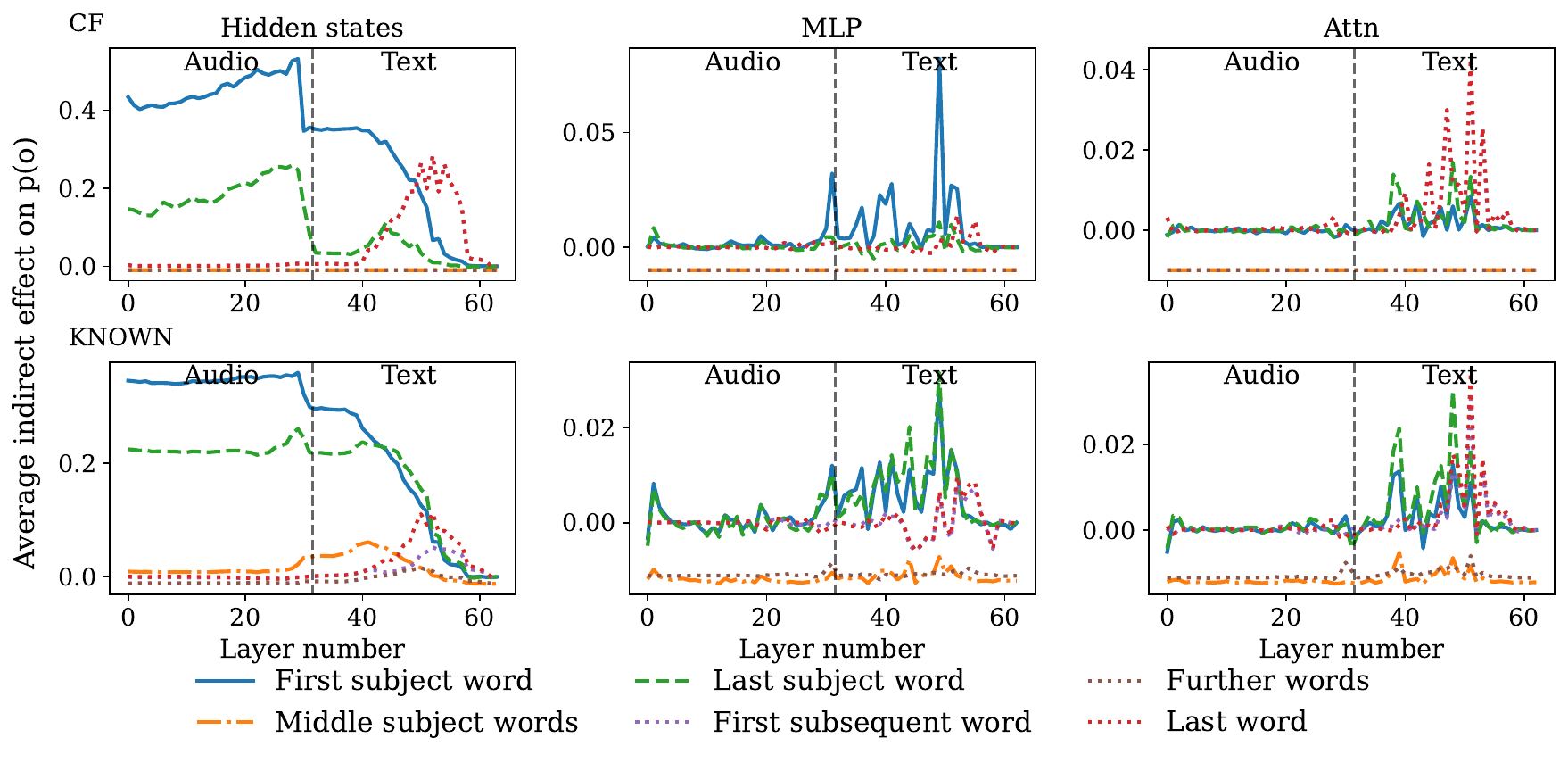}
    \caption{Layer-wise Average Indirect Effect (AIE) of hidden states (left), MLP modules (middle), and attention modules (right) for the CounterFact (top) and Known-1000 (bottom) datasets. The dashed vertical line separates the audio encoder layers (left) from the LLM backbone 
    layers (right).} 
    \vspace{-15pt}
    \label{fig:combined_lineplot}
\end{figure}

\section{Results}
\vspace{-5pt}
\subsection{Causal Tracing}
Figure~\ref{fig:combined_lineplot} reports the layer-wise AIE of hidden states, MLP modules
, and attention modules for CounterFact and Known-1000. Overall, factual information is jointly encoded in audio and text streams, but their contributions differ across layers, modules, and word positions.

For CounterFact, audio-path hidden states show the strongest AIE in middle-to-late encoder layers (around 25–31), while text-path AIE peaks later in the decoder, indicating that the language backbone still supports factual retrieval even with spoken input. Within the subject span, the first subject word carries the largest AIE (solid line), followed by the last subject word (dash line), pointing to a subject-centric storage pattern where early and final audio tokens dominate the causal influence on the predicted object. The last word of the sentence also contributes strongly in the later language backbone 
layers (red line), likely because the model consolidates the full subject–relation–object context there.

MLP modules show that mid-layer text in LLM backbone has the largest AIE, consistent with the view that feed-forward blocks act as key–value memories for factual associations. Attention modules contribute more modest 
but widespread AIE, suggesting a role in routing and integrating information rather than storing facts. 
A similar but less sharply peaked profile is observed on Known-1000, with audio contributions more evenly spread across layers. 

These results indicate that factual knowledge in LALMs is distributed across audio encoder and LLM backbone 
: audio encoder layers primarily initiate retrieval from speech, while LLM backbone 
MLPs consolidate subject-aligned representations into fact memories.

\vspace{-5pt}
\subsection{Performance Comparison of Model Editing}
As shown in Table~\ref{tab:model_edit_roles}, for single-modality editing, targeted editing consistently outperforms fine-tuning in the audio setting, but not for text. This highlights that audio-layer edits play a central role in controlling spoken factual knowledge. Moreover, single-layer editing consistently outperforms multi-layer editing for both audio and text. This is possibly due to that single-layer edits are applied separately for each speech sample and the parameters are reset to their original values between edits, whereas multi-layer editing accumulates updates without resetting, leading to larger distribution shifts that hurt performance.  

We additionally evaluate multi-layer editing with parameter resetting between examples. With resetting, multi-layer editing achieves strong overall performance across all modalities, reaching 78.81 for audio editing, 71.40 for text editing, and 79.67 for multimodal editing.


In the multimodal setting, jointly editing both audio encoder and LLM backbone layers achieves the best overall performance. This indicates that while audio-layer editing serves as the key anchor for modifying spoken factual knowledge, coordinated cross-modal updates further improve factual modification by aligning the updated acoustic representation with the language pathway. 

\begin{table}[t!]
\centering
\caption{Editing performance across modalities and strategies.}
\vspace{-10pt}
\setlength{\tabcolsep}{6pt}
\renewcommand{\arraystretch}{1}
\begin{tabular}{c l c c c c}
\toprule
\textbf{Mod.} & \textbf{Method} & \textbf{Score}$\uparrow$ & \textbf{ES}$\uparrow$ & \textbf{PS}$\uparrow$ & \textbf{NS}$\uparrow$ \\
\midrule
\multicolumn{2}{l}{Baseline}     & 29.09 & 20.4 & 24.4 & 76.18 \\
\midrule
\multirow{3}{*}{Audio}
  & FT           & 61.50 & 97.2 & 45.9 & 59.86 \\
  & Single-layer & 71.36 & 78.2 & 66.5 & \textbf{70.34} \\
  & Multi-layer  &  63.52   &  80.7  &  67.9  &  49.72   \\
\midrule
\multirow{3}{*}{Text}
  & FT           & 68.21 & 97.6 & 70.0 & 51.42 \\
  & Single-layer & 64.10 & 79.7 & 56.3 & 60.64 \\
  & Multi-layer  &  51.05   &  54.02  &  42.2  &  60.18   \\
\midrule
\multirow{3}{*}{Multi}
  & FT           & 68.18 & \textbf{98.40} & 73.10 & 49.60 \\
  & Single-layer & \textbf{77.60} & 95.20 &\textbf{78.70} & 64.72 \\
  & Multi-layer  &  60.91   &  82.5  &  72.0  &  43.02   \\
\bottomrule
\end{tabular}
\vspace{-5pt}
\label{tab:model_edit_roles}
\end{table}


\vspace{-5pt}
\subsection{Layer and Word Selection Analysis}
\begin{table}[t]
\centering
\small
\caption{Editing scores $S$ for Top-3 layers by modality and word position. Layer numbers are ranked in parentheses. }
\vspace{-6pt}
\label{tab:word_layer_selection}
\setlength{\tabcolsep}{6pt}
\renewcommand{\arraystretch}{0.95}
\begin{tabular}{c l c c c}
\toprule
\textbf{Mod.} & \textbf{Word} & \textbf{Top-1} & \textbf{Top-2} & \textbf{Top-3} \\
\midrule
\multirow{2}{*}{Audio}
  & subj.\ first (29,28,27) & 71.70 & 73.34 & 73.34 \\
  & subj.\ last  (29,28,27) & 71.05 & 70.76 & 68.49 \\
\midrule
\multirow{2}{*}{Text}
  & subj.\ first (2,1,7)    & 51.77 & 32.33 & 51.54 \\
  & subj.\ last  (12,10,8)  & 56.47 & 57.63 & 63.88 \\
\bottomrule
\end{tabular}
\vspace{-20pt}
\end{table}







To gain a deep understanding of how causal tracing offers actionable guidance for choosing editing layers, we conduct a word-position–aware layer selection study summarized in Table~\ref{tab:word_layer_selection}. For audio, both subject-first and subject-last words select the same late encoder layers (27–29), but edits anchored on the first word consistently achieve higher overall scores, matching its stronger AIE and indicating that early subject-aligned frames are the most effective targets for modifying spoken facts. For text, subject-first and subject-last words peak at different layers, yet configurations based on the last subject word (layers 8, 10, 12) outperform those based on the first, suggesting that factual associations in the language backbone are most stably formed once the subject phrase is complete. Overall, layers and word positions with higher AIE also yield better, more specific edits, confirming that speech-aware causal tracing provides actionable guidance for choosing editing locations in LALMs.

\section{Conclusion}
We presented a locate–then–edit framework for LALMs that uses speech-aware causal tracing to identify where spoken facts are stored and then applies targeted parameter updates to modify them. Our analysis shows that factual knowledge is jointly encoded across audio and text pathways and that edits guided by causal signals yield more effective and specific updates than standard fine-tuning, especially when coordinating multimodal layers. These results suggest a practical route toward reliable, fine-grained knowledge control in speech-based AI systems.

\section{Generative AI Use Disclosure}
Generative AI tools were used solely for minor language editing and polishing to improve the clarity and readability of the manuscript. These tools were not used to generate scientific content, analyze experimental results, summarize related work, develop methodologies, or propose research ideas. All conceptual contributions, experimental design, analysis, and conclusions presented in this paper were developed and verified by the authors. The authors take full responsibility for the content of this manuscript.

\bibliographystyle{IEEEtran}
\bibliography{mybib}

@misc{alphaedit,
      title={AlphaEdit: Null-Space Constrained Knowledge Editing for Language Models}, 
      author={Junfeng Fang and Houcheng Jiang and Kun Wang and Yunshan Ma and Shi Jie and Xiang Wang and Xiangnan He and Tat-seng Chua},
      year={2025},
      eprint={2410.02355},
      archivePrefix={arXiv},
      primaryClass={cs.CL},
      url={https://arxiv.org/abs/2410.02355}, 
}

@inproceedings{visionLLM,
 author = {Basu, Samyadeep and Grayson, Martin and Morrison, Cecily and Nushi, Besmira and Feizi, Soheil and Massiceti, Daniela},
 booktitle = {Advances in Neural Information Processing Systems},
 doi = {10.52202/079017-0237},
 editor = {A. Globerson and L. Mackey and D. Belgrave and A. Fan and U. Paquet and J. Tomczak and C. Zhang},
 pages = {7400--7426},
 publisher = {Curran Associates, Inc.},
 title = {Understanding Information Storage and Transfer in Multi-Modal Large Language Models},
 url = {https://proceedings.neurips.cc/paper_files/paper/2024/file/0dfe31d6e703e138d46a7d2fced38b7c-Paper-Conference.pdf},
 volume = {37},
 year = {2024}
}

@misc{rome,
      title={Locating and Editing Factual Associations in GPT}, 
      author={Kevin Meng and David Bau and Alex Andonian and Yonatan Belinkov},
      year={2023},
      eprint={2202.05262},
      archivePrefix={arXiv},
      primaryClass={cs.CL},
      url={https://arxiv.org/abs/2202.05262}, 
}

@misc{memit,
      title={Mass-Editing Memory in a Transformer}, 
      author={Kevin Meng and Arnab Sen Sharma and Alex Andonian and Yonatan Belinkov and David Bau},
      year={2023},
      eprint={2210.07229},
      archivePrefix={arXiv},
      primaryClass={cs.CL},
      url={https://arxiv.org/abs/2210.07229}, 
}

@misc{chu2024qwen2audiotechnicalreport,
      title={Qwen2-Audio Technical Report}, 
      author={Yunfei Chu and Jin Xu and Qian Yang and Haojie Wei and Xipin Wei and Zhifang Guo and Yichong Leng and Yuanjun Lv and Jinzheng He and Junyang Lin and Chang Zhou and Jingren Zhou},
      year={2024},
      eprint={2407.10759},
      archivePrefix={arXiv},
      primaryClass={eess.AS},
      url={https://arxiv.org/abs/2407.10759}, 
}

@misc{mlpkeyvalue,
      title={Transformer Feed-Forward Layers Are Key-Value Memories}, 
      author={Mor Geva and Roei Schuster and Jonathan Berant and Omer Levy},
      year={2021},
      eprint={2012.14913},
      archivePrefix={arXiv},
      primaryClass={cs.CL},
      url={https://arxiv.org/abs/2012.14913}, 
}

@misc{yu2023meloenhancingmodelediting,
      title={MELO: Enhancing Model Editing with Neuron-Indexed Dynamic LoRA}, 
      author={Lang Yu and Qin Chen and Jie Zhou and Liang He},
      year={2023},
      eprint={2312.11795},
      archivePrefix={arXiv},
      primaryClass={cs.CL},
      url={https://arxiv.org/abs/2312.11795}, 
}

@inproceedings{zheng-etal-2023-edit,
    title = "Can We Edit Factual Knowledge by In-Context Learning?",
    author = "Zheng, Ce  and
      Li, Lei  and
      Dong, Qingxiu  and
      Fan, Yuxuan  and
      Wu, Zhiyong  and
      Xu, Jingjing  and
      Chang, Baobao",
    editor = "Bouamor, Houda  and
      Pino, Juan  and
      Bali, Kalika",
    booktitle = "Proceedings of the 2023 Conference on Empirical Methods in Natural Language Processing",
    month = dec,
    year = "2023",
    address = "Singapore",
    publisher = "Association for Computational Linguistics",
    url = "https://aclanthology.org/2023.emnlp-main.296/",
    doi = "10.18653/v1/2023.emnlp-main.296",
    pages = "4862--4876"
}

@inproceedings{whisperx,
  author = {Bain, Max and Huh, Jaesung and Han, Tengda and Zisserman, Andrew},
  title = {WhisperX: Time-Accurate Speech Transcription of Long-Form Audio},
  booktitle = {Proc. {INTERSPEECH} 2023},
  year = {2023}
}

@misc{blsp,
      title={BLSP: Bootstrapping Language-Speech Pre-training via Behavior Alignment of Continuation Writing},
      author={Chen Wang and Minpeng Liao and Zhongqiang Huang and Jinliang Lu and Junhong Wu and Yuchen Liu and Chengqing Zong and Jiajun Zhang},
      year={2023},
      eprint={2309.00916},
      archivePrefix={arXiv},
      primaryClass={cs.CL},
      url={https://arxiv.org/abs/2309.00916},
}

@misc{tang2024salmonngenerichearingabilities,
      title={SALMONN: Towards Generic Hearing Abilities for Large Language Models}, 
      author={Changli Tang and Wenyi Yu and Guangzhi Sun and Xianzhao Chen and Tian Tan and Wei Li and Lu Lu and Zejun Ma and Chao Zhang},
      year={2024},
      eprint={2310.13289},
      archivePrefix={arXiv},
      primaryClass={cs.SD},
      url={https://arxiv.org/abs/2310.13289}, 
}

@misc{gemini25_tts,
  author       = {{Google Cloud}},
  title        = {Gemini 2.5 Flash TTS},
  year         = {2025},
  howpublished = {\url{https://cloud.google.com/text-to-speech/docs/gemini-tts}},
  note         = {Accessed 2026}
}

@inproceedings{panayotov2015librispeech,
  title={Librispeech: an ASR corpus based on public domain audio books},
  author={Panayotov, Vassil and Chen, Guoguo and Povey, Daniel and Khudanpur, Sanjeev},
  booktitle={Acoustics, Speech and Signal Processing (ICASSP), 2015 IEEE International Conference on},
  pages={5206--5210},
  year={2015},
  organization={IEEE}
}

@article{DBLP:journals/corr/abs-2005-11401,
  author       = {Patrick Lewis and
                  Ethan Perez and
                  Aleksandra Piktus and
                  Fabio Petroni and
                  Vladimir Karpukhin and
                  Naman Goyal and
                  Heinrich K{\"{u}}ttler and
                  Mike Lewis and
                  Wen{-}tau Yih and
                  Tim Rockt{\"{a}}schel and
                  Sebastian Riedel and
                  Douwe Kiela},
  title        = {Retrieval-Augmented Generation for Knowledge-Intensive {NLP} Tasks},
  journal      = {CoRR},
  volume       = {abs/2005.11401},
  year         = {2020},
  url          = {https://arxiv.org/abs/2005.11401},
  eprinttype    = {arXiv},
  eprint       = {2005.11401},
  bibsource    = {dblp computer science bibliography, https://dblp.org}
}

@article{DBLP:journals/corr/abs-1911-00172,
  author       = {Urvashi Khandelwal and
                  Omer Levy and
                  Dan Jurafsky and
                  Luke Zettlemoyer and
                  Mike Lewis},
  title        = {Generalization through Memorization: Nearest Neighbor Language Models},
  journal      = {CoRR},
  volume       = {abs/1911.00172},
  year         = {2019},
  url          = {http://arxiv.org/abs/1911.00172},
  eprinttype    = {arXiv},
  eprint       = {1911.00172},
  bibsource    = {dblp computer science bibliography, https://dblp.org}
}

@article{goel2025audio,
  title={Audio flamingo 3: Advancing audio intelligence with fully open large audio language models},
  author={Goel, Arushi and Ghosh, Sreyan and Kim, Jaehyeon and Kumar, Sonal and Kong, Zhifeng and Lee, Sang-gil and Yang, Chao-Han Huck and Duraiswami, Ramani and Manocha, Dinesh and Valle, Rafael and others},
  journal={arXiv preprint arXiv:2507.08128},
  year={2025}
}

@inproceedings{ghosh-etal-2024-gama,
    title = "{GAMA}: A Large Audio-Language Model with Advanced Audio Understanding and Complex Reasoning Abilities",
    author = "Ghosh, Sreyan  and
      Kumar, Sonal  and
      Seth, Ashish  and
      Evuru, Chandra Kiran Reddy  and
      Tyagi, Utkarsh  and
      Sakshi, S  and
      Nieto, Oriol  and
      Duraiswami, Ramani  and
      Manocha, Dinesh",
    editor = "Al-Onaizan, Yaser  and
      Bansal, Mohit  and
      Chen, Yun-Nung",
    booktitle = "Proceedings of the 2024 Conference on Empirical Methods in Natural Language Processing",
    month = nov,
    year = "2024",
    address = "Miami, Florida, USA",
    publisher = "Association for Computational Linguistics",
    url = "https://aclanthology.org/2024.emnlp-main.361/",
    doi = "10.18653/v1/2024.emnlp-main.361",
    pages = "6288--6313"
}

@article{Das2024SpeechVerseAL,
  title={SpeechVerse: A Large-scale Generalizable Audio Language Model},
  author={Nilaksh Das and Saket Dingliwal and S. Ronanki and Rohit Paturi and David Huang and Prashant Mathur and Jie Yuan and Dhanush Bekal and Xing Niu and Sai Muralidhar Jayanthi and Xilai Li and Karel Mundnich and Monica Sunkara and Sundararajan Srinivasan and Kyu J Han and Katrin Kirchhoff},
  journal={ArXiv},
  year={2024},
  volume={abs/2405.08295},
  url={https://api.semanticscholar.org/CorpusID:269761552}
}

@misc{leng2024cursemultimodalitiesevaluatinghallucinations,
      title={The Curse of Multi-Modalities: Evaluating Hallucinations of Large Multimodal Models across Language, Visual, and Audio}, 
      author={Sicong Leng and Yun Xing and Zesen Cheng and Yang Zhou and Hang Zhang and Xin Li and Deli Zhao and Shijian Lu and Chunyan Miao and Lidong Bing},
      year={2024},
      eprint={2410.12787},
      archivePrefix={arXiv},
      primaryClass={cs.CV},
      url={https://arxiv.org/abs/2410.12787}, 
}

@misc{kuan2024understandingsoundsmissingquestions,
      title={Understanding Sounds, Missing the Questions: The Challenge of Object Hallucination in Large Audio-Language Models}, 
      author={Chun-Yi Kuan and Wei-Ping Huang and Hung-yi Lee},
      year={2024},
      eprint={2406.08402},
      archivePrefix={arXiv},
      primaryClass={eess.AS},
      url={https://arxiv.org/abs/2406.08402}, 
}

@misc{kuan2024largeaudiolanguagemodelstruly,
      title={Can Large Audio-Language Models Truly Hear? Tackling Hallucinations with Multi-Task Assessment and Stepwise Audio Reasoning}, 
      author={Chun-Yi Kuan and Hung-yi Lee},
      year={2024},
      eprint={2410.16130},
      archivePrefix={arXiv},
      primaryClass={eess.AS},
      url={https://arxiv.org/abs/2410.16130}, 
}

@inproceedings{yue2023token2vec,
    author       = {{Xianghu Yue and Junyi Ao and Xiaoxue Gao and Haizhou Li}},
    title        = {{Token2vec: A Joint Self-Supervised Pre-Training Framework Using Unpaired Speech and Text}},
    booktitle    = {Proc. {ICASSP} 2023 -- IEEE International Conference on Acoustics, Speech and Signal Processing},
    address      = {{Rhodes Island, Greece}},
    month        = {{Jun.}},
    year         = 2023,
    publisher    = {{IEEE}},
    doi          = {10.1109/ICASSP49357.2023.10096923}
}

\end{document}